# Benchmarking Motivational Interviewing Competence of Large Language Models

Aishwariya Jha [a], Prakrithi Shivaprakash [a], Lekhansh Shukla [a], Animesh Mukherjee [b], Prabhat Chand [a], Pratima Murthy [c].

[a] Centre for Addiction Medicine, Department of Psychiatry, National Institute of Mental Health and Neuro Sciences (NIMHANS), Bengaluru, India.

[b] Department of Computer Science and Engineering, Indian Institute of Technology (IIT) Kharagpur, India.

[c] Department of Psychiatry, National Institute of Mental Health and Neuro Sciences (NIMHANS), Bengaluru, India.

**Corresponding author:**

Dr Lekhansh Shukla

E-mail: drlekhansh@gmail.com

Address: Office of the Centre for Addiction Medicine (CAM), 2nd Floor, CAM ward – Female Wing, National Institute of Mental Health and Neuro Sciences (NIMHANS), Hosur Road, Bengaluru, India.

Telephone: +91 9886160956

# Abstract

**Introduction:** Motivational interviewing (MI) promotes behavioural change in substance use disorders. Its fidelity is measured using the Motivational Interviewing Treatment Integrity (MITI) framework. While large language models (LLMs) can potentially generate MI-consistent therapist responses, their competence using MITI is not well-researched, especially in real-world clinical transcripts. We aim to benchmark MI competence of proprietary and open-source models compared to human therapists in real-world transcripts and assess distinguishability from human therapists.

**Methods:** We shortlisted 3 proprietary and 7 open-source LLMs from LMArena, evaluated performance using MITI 4.2 framework on two datasets (96 handcrafted model transcripts, 34 real-world clinical transcripts). We generated parallel LLM-therapist utterances iteratively for each transcript while keeping client responses static, and ranked performance using a composite ranking system with MITI components and verbosity. We conducted a distinguishability experiment with two independent psychiatrists to identify human-vs-LLM responses.

**Results:** All 10 tested LLMs had fair (MITI global scores >3.5) to good (MITI global scores >4) competence across MITI measures, and three best-performing models (gemma-3-27b-it, gemini-2.5-pro, grok-3) were tested on real-world transcripts. All showed good competence, with LLMs outperforming human-expert in Complex Reflection percentage (39% vs 96%) and Reflection-Question ratio (1.2 vs >2.8). In the distinguishability experiment, psychiatrists identified LLM responses with only 56% accuracy, with d-prime: 0.17 and 0.25 for gemini-2.5-pro and gemma-3-27b-it respectively.

**Conclusion:** LLMs can achieve good MI proficiency in real-world clinical transcripts using MITI framework. These findings suggest that even open-source LLMs are viable candidates for expanding MI counselling sessions in low-resource settings, warranting further clinical validation.

**Keywords:** Motivational interviewing, Substance-related disorders, Large language models, Benchmarking, Artificial intelligence

# 1  Background

Motivational Interviewing (MI) is a collaborative, client-centred counselling technique to promote behavioural change ([Miller & Rollnick, 2013](#)). Briefly, in MI, the therapist tries to evoke clients' reasons for change using reflections and open-ended questions. The focus is on affirming client strengths, emphasising their autonomy, and seeking collaboration, while avoiding behaviours that can provoke defensiveness, such as confrontation and persuasion. Since its origin in the 1980s, MI has been shown to be better than no intervention or feedback for alcohol, tobacco and cannabis use ([Schwenker et al., 2023](#)). However, there have been two challenges in implementing and studying MI at scale. First, it is a difficult technique to imbibe and requires ongoing supervision to maintain fidelity ([Hall et al., 2016](#)). *Rollnick et al.* have indicated that even with training and supervision, some practitioners may not be able to practice the spirit of MI ([Rollnick & Gobat, 2016](#)). Second, most effectiveness studies have not measured treatment fidelity, and thus, we do not know what actually worked ([Schwenker et al., 2023](#)). In this context, we discuss attempts to develop computational solutions that can deliver MI.

We restrict our discussion to generative Artificial Intelligence (AI) related work, specifically Large Language Models (LLMs). LLMs are deep learning models which are designed to understand and generate human-like text. These models are available in two varieties: pretrained language models, which have only been trained to predict the next word in a sequence, and instruction-tuned models, which have received additional training to follow natural language instructions ([Zhao et al., 2023](#)). *Karve et al.* have reviewed the extant research on LLMs for MI ([Karve et al., 2025](#)), and the early signals are encouraging. For example, all fifteen reviewed studies reported good feasibility and acceptability. We must consider the presumptions made in these studies and the limitations of their study designs.

First, most studies have tested the deployment of a chosen LLM for conducting online MI for various indications and measured user satisfaction or effectiveness. This design presumes substantial MI competence in LLMs. Only 6 out of 15 studies used any fidelity check ([Karve et al., 2025](#)). Even when used, 'expert-rating' is chosen over Motivational Interviewing Treatment Integrity (MITI) coding ([Moyers et al., 2016](#)). MITI scoring is more informative than a global 'expert rating' as it provides technical and relational competence. It also has a standardised threshold for MI competence. A subset of studies has measured the ability of a chosen LLM to produce a specific component of MI, like reflections or open-ended questions. Here again, it is important to realise that MI is a gestalt of many techniques and needs holistic assessment.

Second, none of the studies have systematically compared the competence of LLMs with that of addiction psychiatry experts. Therefore, we do not know if LLMs are as good as the current standard of care for a given target behaviour. This comparison should ideally be done in two settings. One is an artificial setting in which the expert and LLMs generate therapist utterances for a given MI session. This allows sufficient time for the expert to generate the best response, unlike in real-life situations, where a certain cadence of dialogue must be maintained. Another is a realistic setting, in which LLMs generate therapist utterances for a recorded and transcribed clinical MI session. These two settings answer distinct questions: how good are LLMs as compared to the usual performance of addiction specialists, and how good are LLMs as compared to the best performance of addiction specialists?

Finally, there are no studies which have examined the distinguishability of LLM-generated therapist utterances from those of addiction specialists. Distinguishability is a proxy for the human-like quality of interactions and is therefore important.

With this background, we aim to: (i) Measure MI competence of proprietary and open-source LLMs using MITI-4.2. (ii) Compare the performance of the top open-source and proprietary LLMs with that of an addiction specialist. (iii) Assess distinguishability between human-delivered and AI-generated MI responses using two independent addiction psychiatrists.

# 2 Methods

This study was conducted in an addiction medicine unit of a tertiary-level teaching hospital in South India between January and December 2025. The study received ethical approval from the Institutional Ethics Committee. Figure 1 shows the study design.

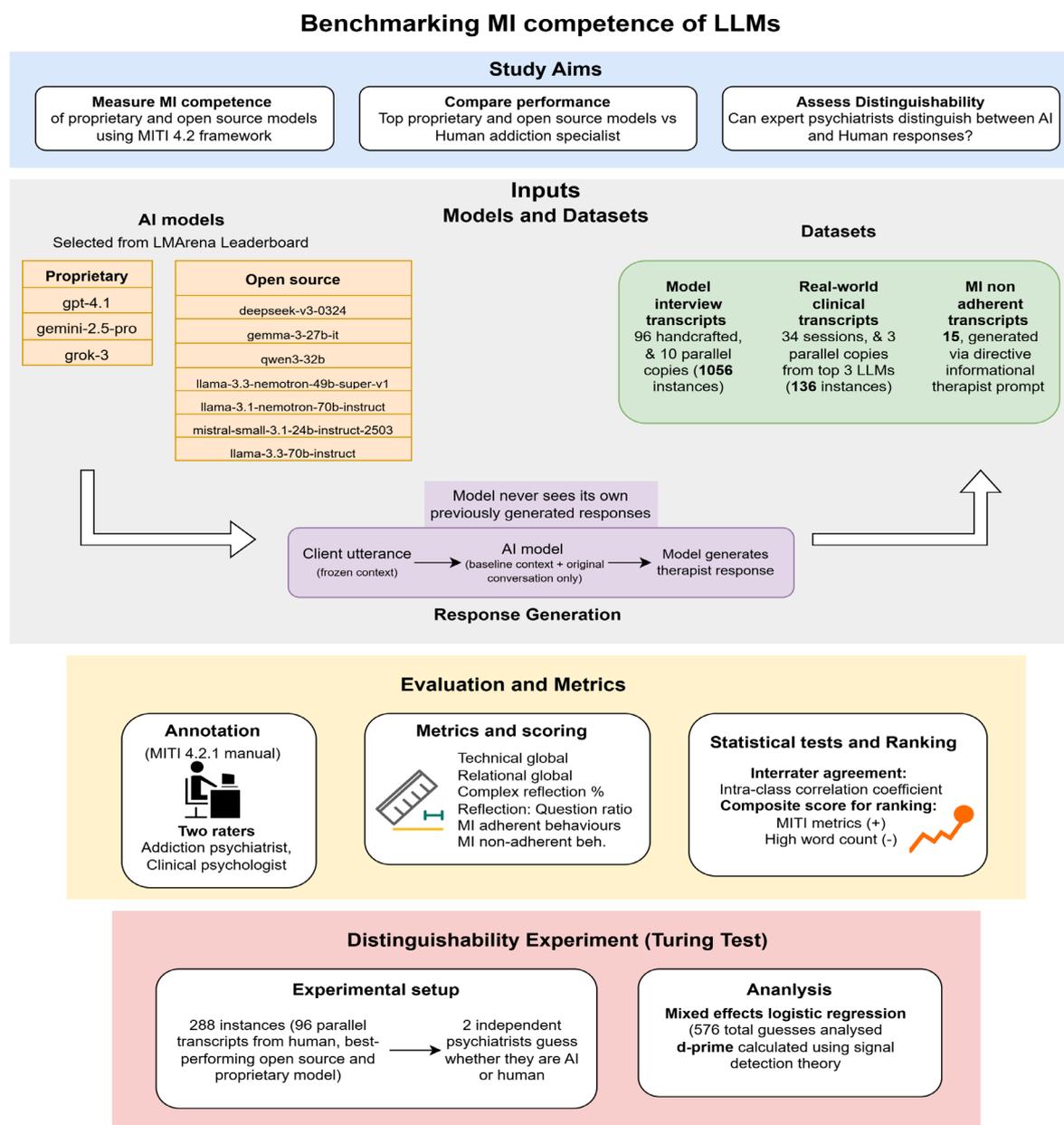

*Figure 1: Study design*

## 2.1 Model shortlisting

Models were shortlisted using the *LMArena* leaderboard (https://lmarena.ai/leaderboard) as of 31st May 2025. We chose three proprietary models: gpt-4.1, gemini-2.5-pro (Comanici et al., 2025), grok-3, and seven open-source models (deepseek-v3-0324, gemma-3-27b-it, qwen3-32b, llama-3.3-nemotron-49b-super-v1, llama-3.1-nemotron-70b-instruct, mistral-small-3.1-24b-instruct-2503 and llama-3.3-70b-instruct). Supplementary Table S1 describes the filtering criteria used for selection.

## 2.2 LLM-therapist response generation

Models were to be evaluated on their ability to generate MI-consistent responses to client utterances. The model's task, therefore, was to generate one therapist utterance conditioned on the previous conversation turns. At each turn, the model only sees client and therapist utterances made by the investigators, and never its own previous utterances. The rationale for this design choice is to ensure that the text presented to it remains coherent. The algorithm is shown in Figure 2 (see code in Supplementary Box S1).

We iteratively built a detailed prompt to optimise MI adherence by testing increasingly detailed prompts. The final prompt is shared in Supplementary Box S2. All responses were obtained using the respective model's default generation configuration via API calls to Open Router.

# How are AI-Therapist responses generated?

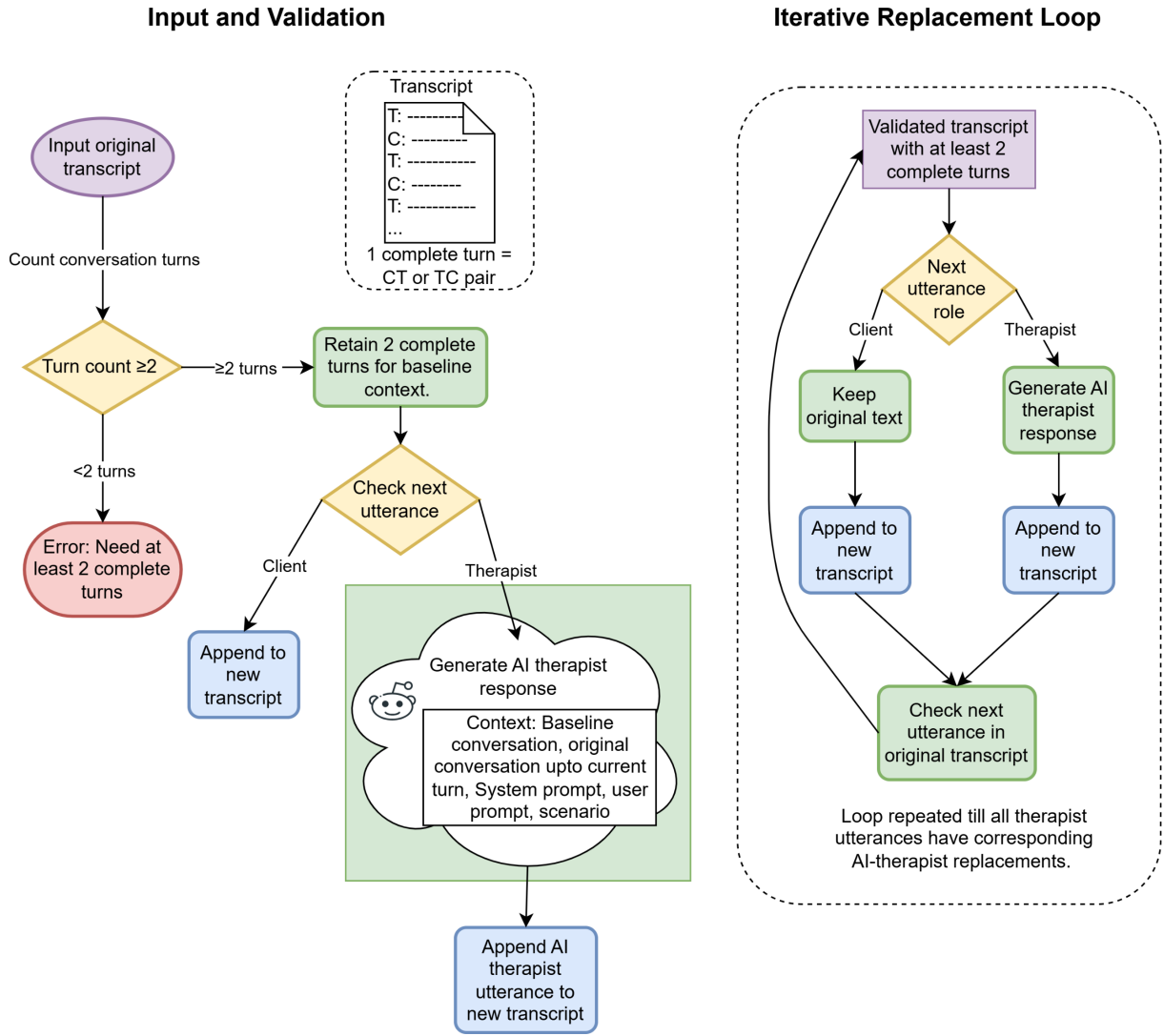

*Figure 2: Algorithm for generating LLM-therapist responses for evaluation of MI skills*

## 2.3 Dataset Construction

We collected three types of datasets for this work

**Model Interview Transcripts:** A set of 100 transcripts using MI training educational videos (N=55) ([psychotherapy.net, 2026](psychotherapy.net, 2026); [The Change Companies, 2023](The Change Companies, 2023)) and recollection of client interactions (N=45). These transcripts were handcrafted to provide sufficient opportunities for practising MI. Three independent addiction psychiatrists reviewed these and excluded four transcripts based on the feedback. Finally, an addiction psychiatrist (AJ) rewrote the therapist utterances to the best of her ability to maximise MI adherence.

For each instance of these 96 transcripts, we have 10 copies, one for each model under study. The final size of this dataset is thus 96 * 11 = 1056.

**Real-world clinical interview transcripts:** Adult patients seeking help for substance use were recruited with written informed consent (N = 30). MI sessions conducted by an addiction psychiatrist (N=34) were audio-recorded, and recordings were transcribed (translated if not in English) and anonymised.

We selected three LLMs (one open-source, two proprietary) that had achieved the best performance on model transcripts for this comparison. As a result, we have three copies for each instance of these 34 transcripts. The final size of this dataset is thus 34 * 4 = 136.

**MI non-adherent transcripts:** We sampled 15 model transcripts and prompted Gemini-2.5-pro to play the role of a directive-informational therapist. This was done to obtain some representation of MI non-adherent behaviours, such as confrontation, for the interrater agreement exercise. The prompt is given in Supplementary Box S3.

## 2.4 Annotation setup

We used the MITI 4.2.1 coding manual for all annotations ([Moyers et al., 2014](#)). The MITI 4.2 metrics consist of Technical Global Scores, Relational Global Scores, Complex Reflection (CR) percentage, Reflection: Question (RQ) ratio, and Behavioural Counts, including MI-adherent, MI-non-adherent, and MI-neutral behaviours (Supplementary Figure S1).

An addiction psychiatrist (AJ) and a clinical psychologist learnt MITI coding using the manual and twenty example transcripts (https://casaa.unm.edu/tools/miti.html). Following this, for three rounds, they coded 10 randomly selected transcripts, and disagreements were resolved by discussion. Finally, a dataset of 100 transcripts was constructed by combining 15 MI non-adherent transcripts and 85 randomly selected transcripts from the pool of 1026 model transcripts that the annotators had not yet seen. We used the intraclass correlation coefficient (two-way random effects) to assess interrater agreement.

The therapist's identity was concealed from the annotators.

## 2.5 Measuring and comparing MI competence of LLMs

We report MI competence of LLMs using the following MITI metrics: Relational global scores (Fair=3.5, Good=4), Technical global scores (Fair=3, Good=4), CR percentage (Fair=40%, Good=50%), and RQ ratio (Fair=1, Good=2).

For comparison, we use a ranking procedure. Each transcript (96 model transcripts and 34 real-world transcripts) is considered a test taken by 11 therapist agents (10 LLMs, 1 human) in the case of model transcripts or 4 therapist agents (3 LLMs, 1 human) in the case of real-world transcripts. They get a composite score based on the above four MITI metrics as positively coded variables and the median number of words per therapist utterance as a negatively coded variable to penalise verbosity (Supplementary Box S4). We then rank the performance of all agents for a given transcript. Finally, we compare the rank distributions of all agents to identify those that consistently outperform others.

## 2.6 Distinguishability experiment

In addition to comparing model performance solely on MITI metrics, we also examined whether LLM-therapist responses were distinguishable from human-therapist responses. For this, we used the copies of 96 model transcripts – one where therapist responses are written by the addiction psychiatrist and one each where therapist responses are written by the best performing proprietary and open source LLM (total: 96 * 3 = 288).

These transcripts were presented to two independent addiction psychiatrists in random order, with the task of guessing whether therapist utterances were human or LLM. Raters were blind to each other's decisions. They were informed that they could expect that two-thirds of instances would involve an LLM playing the role of a therapist. Thus, the baseline expectancy of AI is 67%.

We have analysed the resulting (288 * 2 = 576) decisions using a mixed-effects, logistic regression model with random effects for instance (288) and fixed effects for the identity of therapist agent (3; human, LLM1, LLM2) and for the rater (2). Further, we use the signal detection theory and compute *d-prime,* which is a measure of signal strength (distinctiveness of LLM responses in this case) relative to the noise ([Macmillan & Creelman, 2005](#)).

# 3 Results

A comparison of model transcripts and real-world transcripts shows that the latter had more conversation turns, but each turn was shorter. Table 1 shows a comparison between the two datasets.

*Table 1: Datasets used for measuring the competence of large language models in conducting motivational interviewing*

|   | **Characteristic** | **Model Transcripts** | **Real-world Transcripts** | **Statistical Test** |
|---|---|---|---|---|
| 1 | Target Behaviour; N (%) | | | |
|   | - Alcohol or Tobacco | 96 (100 %) | 20 (59 %) | $X^2 = 41$, $p < 0.001$ |
|   | - Other substances | 0 (0 %) | 14 (41 %) | - |
| 2 | Therapist Turns; Median (IQR) | 12.0 (8.2-15.0) | 63.5 (52.5-98.5) | U=62, p<0.001 |
| 3 | Client Turns; Median (IQR) | 11.0 (8.0-14.0) | 62.5 (52.2-98.0) | U=54, p<0.001 |
| 4 | Words per turn, therapist; Median (IQR) | 19.6 (15.7-24.4) | 10.6 (8.5-15.3) | U=2699, p<0.001 |
| 5 | Words per turn, client; Median (IQR) | 24.6 (22.9-28.3) | 19.3 (12.6-29.6) | U=2163, p=0.010 |
| 6 | Client: Therapist Word ratio; Median (IQR) | 1.3 (1.0-1.7) | 1.8 (1.1-2.5) | U=1259, p=0.034 |

The clinical details of patients who contributed to real-world transcripts are given in Supplementary Table S2.

There was good to excellent agreement among the annotators, except for rarer codes such as seeking collaboration. Supplementary Table S3 reports the intra-class correlation (ICC) analysis results, and Supplementary Figure S2 shows within-1 agreement for global scores.

We first report MI competence of LLMs using model and real-world transcripts, and then report the findings of the distinguishability experiment.

## 3.1 Benchmarking MI competence of LLMs

### 3.1.1 MI competence in model transcripts

We see that all LLMs show 'good competence' in all measures except the RQ ratio (Figure 3). Only gemini-pro-2.5 and gemma-3-27b-it have a median RQ ratio higher than 2. Nevertheless, except for two open-source models, all tested LLMs exhibit 'fair' or 'good' competence across all four measures.

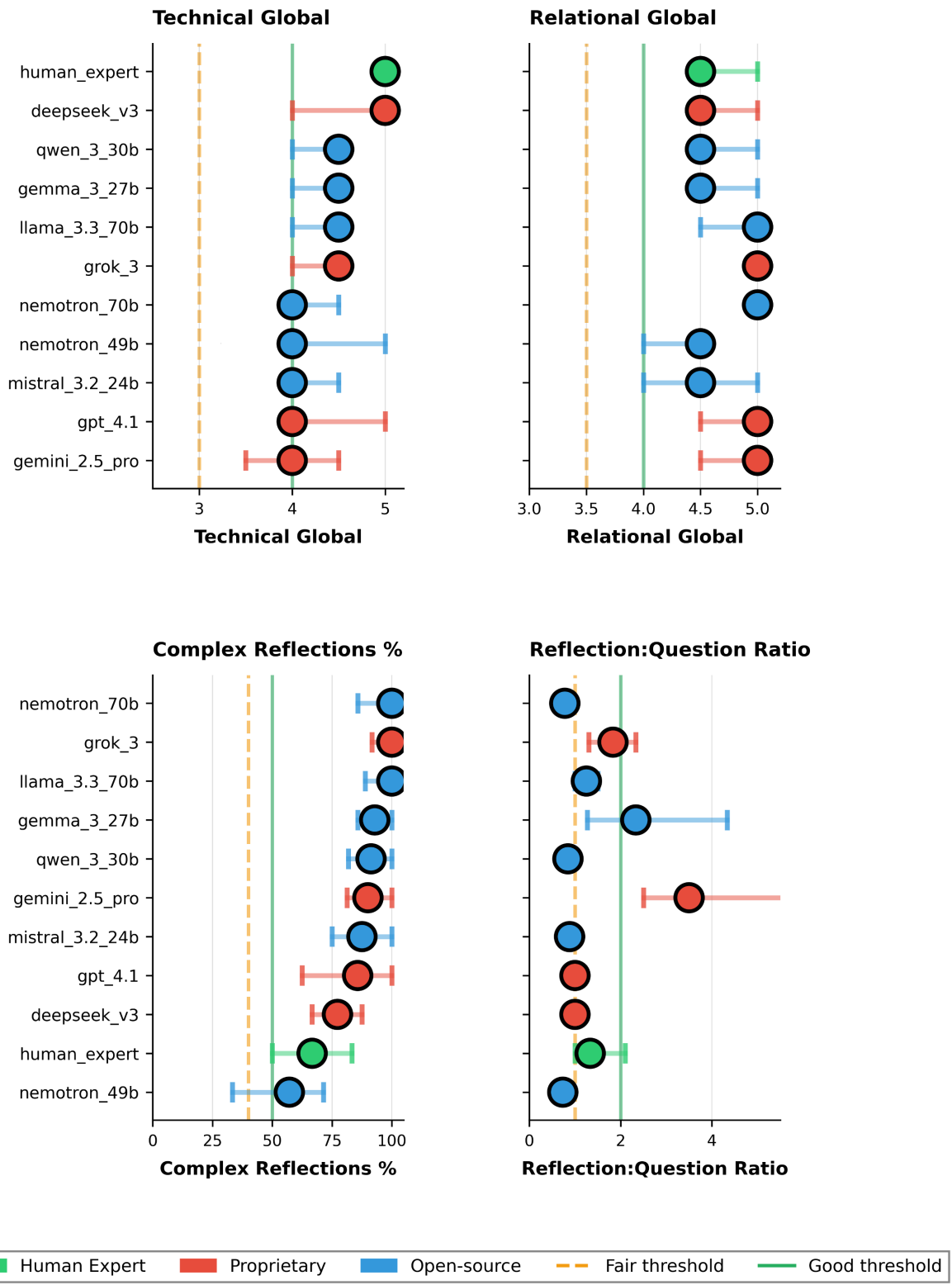

Figure 3: MI competence of Large Language Models using four key MITI metrics and 96 model MI transcripts

We confirmed these findings by ranking each model on each transcript and examining the distribution of these ranks (See Supplementary Box S4 for detailed ranking procedure). With this ranking system, at the time of evaluation (December 2025), grok-3 and gemini-2.5-pro were the most competitive models for MI, at times outperforming the human expert. Gemma-3-27b-it was the most competitive open-source model. Figure 4 shows the final ordinal ranks.

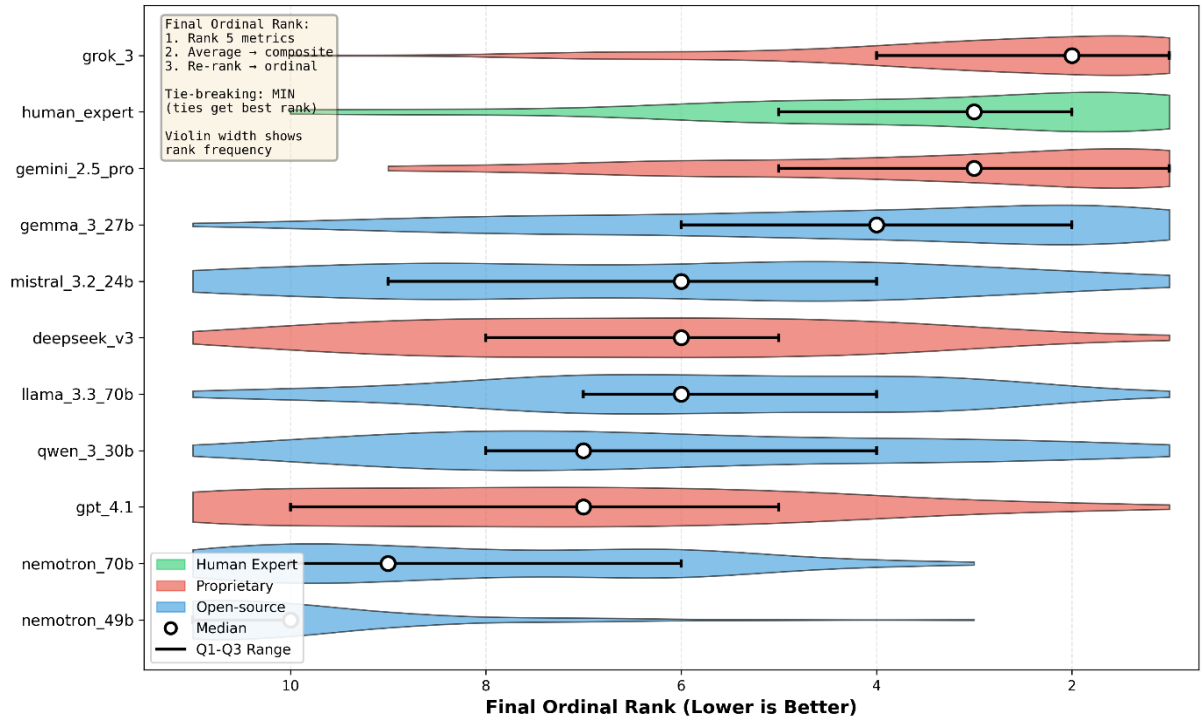

*Figure 4: Performance of 10 Large Language Models and an addiction psychiatrist in conducting motivational interviewing, measured using 96 model transcripts*

Finally, we also evaluated another important aspect of MI – the presence of MI-adherent behaviours and the absence of MI-non-adherent behaviours (See Supplementary Figure S1 for the list). While the count of MI-non-adherent behaviours was zero across all models, the addiction psychiatrist, followed by grok-3, demonstrated the highest MI-adherent behaviour count per conversation. LLMs produced more reflections and questions as compared to a human therapist (Supplementary Figure S3).

### 3.1.2 MI competence in real-world clinical MI transcripts

LLM-therapist responses were obtained from gemma-3-27b-it, gemini-2.5-pro, and grok-3 (chosen from above), resulting in a total of 136 transcripts (34 original and 3 parallel LLM versions), all of which were MITI-coded.

We observe that LLMs maintain good competence even on real-world transcripts across all four measures used in MITI 4.2 (Figure 5). Furthermore, using the ranking methodology, we find that they substantially outperform the addiction psychiatrist. Of the 34 transcripts, each a test, gemini-pro-2.5 ranked first 26 times, whereas the addiction psychiatrist ranked first only 7 times. Supplementary Figure S4 shows the distribution of ranks.

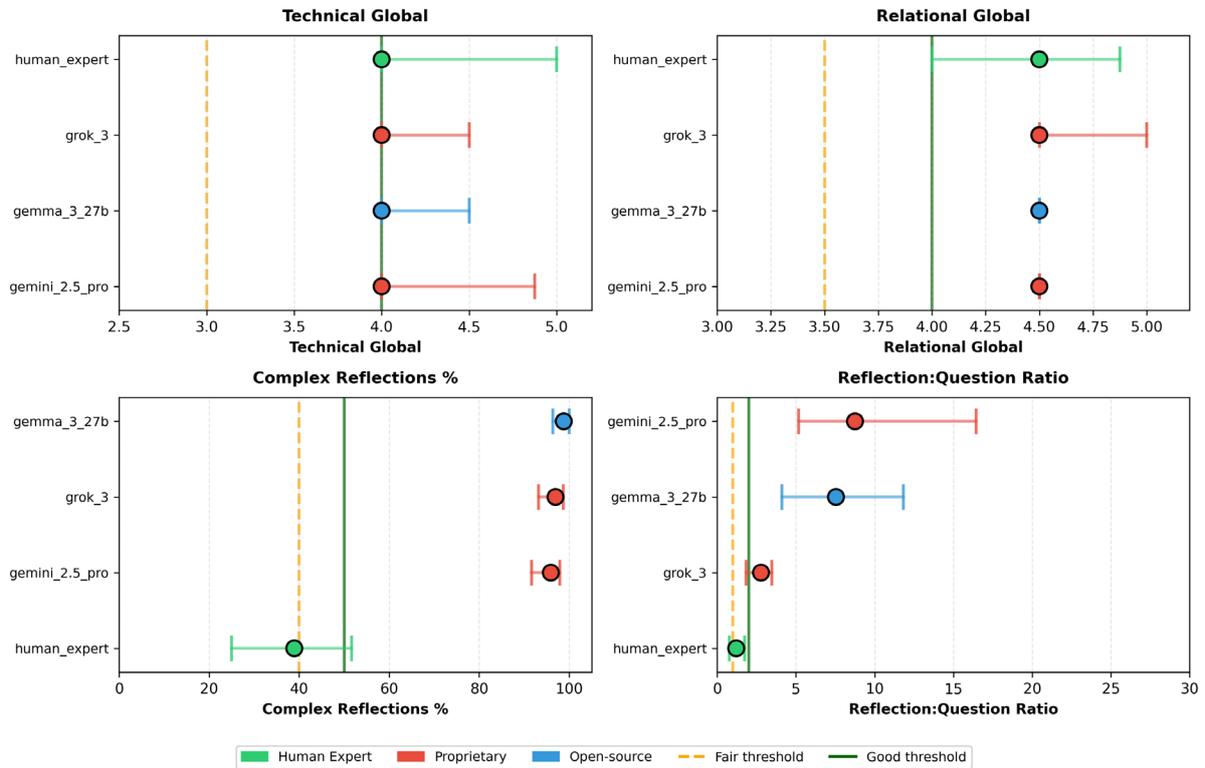

*Figure 5: MI competence of Large Language Models using four key MITI metrics and 34 real-world MI transcripts*

However, LLM responses were significantly longer than human, with median words per utterance of 7, 27, 45.8, and 45 for human-expert, gemini-2.5-pro, grok-3, and gemma-3-27b-it, respectively. See Figure 6 and Supplementary Table S4 for a side-by-side comparison of a LLM therapist's responses compared to the responses of an addiction psychiatrist.

**LLM-therapist and human-expert responses in a real-world clinical transcript snippet**

| | | | | |
|---|---|---|---|---|
| C: My father was an alcoholic. He was also a heavy smoker. He passed away about 23 years ago. | | | | |
| T: Did your father being an alcoholic affect you? | | | | |
| C: Yes.Yes.And the worst part was that my mom was in Australia at that time, because my sister was about to give birth to her second child. I had gone to work. One day, on the way back home, I had forgotten something. I opened the door and saw my dad drunk. It was early morning, around 9:30. He couldn't even stand. What can you say? What can you do? | | | | |
| T: How did it influence you, seeing your father drinking early in the morning? | | | | |
| C: He couldn't even talk properly. I had to hold him and put him on the bed. His lips were drooping. I felt like punching him, but, what can you do? He is your father. I couldn't do much. | | | | |
| You didn't feel good about it? | That was a deeply upsetting and frustrating experience for you, to see your father in that state and feel like there was nothing you could do about it. | That sounds like a really difficult and upsetting experience, witnessing that and feeling so helpless. It's understandable you'd have such strong, mixed feelings in that moment – anger, frustration, and yet still a sense of responsibility towards your father. | It sounds like that moment was incredibly painful for you, carrying both anger and helplessness. I can see how much it affected you to witness that. | |
| Of course. I mean, drinking has a limit and a time. If it's 7 or 8 in the evening, fine. But not in the morning—8:30 or 9:00. Come on. And he would drink straight. | | | | |
| No dilution? | That experience with your father really drew a clear line for you about drinking, and what he was doing was way over that line. | It sounds like witnessing that was incredibly unsettling, especially knowing your mother was away and you felt solely responsible in that moment. It's understandable to feel a mix of anger and helplessness seeing someone you care about struggle like that. | It sounds like seeing your father drink at such an inappropriate time really upset you, and it's clear you have strong feelings about when and how much drinking is acceptable. | |
| No, nothing. He would just gulp about one quarter. | | | | |
| You didn't want to do that yourself? | That's a powerful and disturbing image to carry with you. Seeing him like that, it crossed a line for you, a fundamental rule about when and how a person should drink. | It sounds like witnessing that was incredibly distressing, and left you feeling helpless and even angry, caught between wanting to help and feeling frustrated with the situation. | It sounds like seeing your father drink like that, especially so early in the day, really troubled you and left a strong impression. | |
| No. Until I finished my graduation, I never smoked and never drank alcohol. I was more into sports. | | | | |
| Okay. | So you made a very conscious choice to go in a completely different direction from your father, focusing on sports instead of alcohol. | So, despite witnessing your father's struggles with alcohol, you intentionally chose a different path for yourself, focusing on sports and staying away from both smoking and drinking until after graduation—that sounds like a real commitment to your health and well-being. | It sounds like you made a conscious choice to steer clear of drinking during those years, focusing on sports instead. I really respect that determination. | |

Legend:
Client and various therapist utterances — Client | Human therapist | Gemini-2.5-pro | Gemma-3-27b-it | Grok-3

*Figure 6: Qualitative example of LLM-therapist and human-expert responses in a real-world clinical transcript*

## 3.2 Distinguishability experiment

While a higher proportion of LLM-therapist transcripts were identified as AI-generated (0.57 for gemini-2.5-pro and 0.60 for gemma-3-27b-it) than human-therapist transcripts (0.51), this was not statistically significant.

Table 2 shows the analysis predicting the likelihood that a transcript will be marked as AI.

*Table 2: Mixed-effects logistic regression predicting the likelihood of a transcript being marked as AI (N=576)*

| Predictor (Fixed effects only) | Odds ratio | Standard error | Z statistic | p value | 95% confidence interval |
|---|---|---|---|---|---|
| Intercept (Human transcript) | 1.13 | 0.20 | 0.67 | 0.50 | 0.79 - 1.61 |
| Therapist: Gemini | 1.34 | 0.30 | 1.29 | 0.20 | 0.86 – 2.08 |
| Therapist: Gemma | 1.53 | 0.35 | 1.87 | 0.06 | 0.98 - 2.38 |

The overall accuracy in identifying AI responses is 0.56, which is lower than the chance accuracy of 0.67, given that two-thirds of the transcripts had an LLM playing the role of a therapist. There was poor agreement between the two experts (kappa = 0.11, p = 0.07).

Signal detection analysis also confirms that LLM responses are almost indistinguishable from human responses, with d-prime values less than 1 for both models (Supplementary Table S5).

# 4 Discussion

We set out to measure the MI competence of LLMs using methods recommended for human therapists, compare their competence with addiction specialists and assess the distinguishability of LLMs playing the role of MI therapists. Before discussing our findings, we wish to discuss some epistemic issues inherent to such inquiries.

First, we presume (with good evidence) that treatment fidelity is related to treatment efficacy and thus measuring it is the first step. For example, therapists' MI adherent behaviours are causally linked to clients' change talk, which is in turn a predictor of clinical improvement (Magill et al., 2014). We also aver that acceptability ratings from users, while important, are not necessarily linked to treatment fidelity or clinical improvement.

Second, the exact methodology by which LLMs are made to take on the role of a therapist is an important design choice. A therapeutic dialogue requires two sentient agents and a mode of communication. One option is to use an LLM to simulate clients; however, earlier studies have shown that LLMs tend to be overly compliant and unrealistically agreeable (Kiuchi et al., 2025; Steenstra et al., 2024). Another choice is to use volunteers who interact in a chatbot interface. For example, the MIBot dataset (Mahmood et al., 2025) has 106 conversations between MIBot and human smokers. In this study, we use an offline method to generate LLM responses, in which the client utterances are frozen. With these design issues in mind, we discuss the main findings of our study.

We found that most LLMs exhibit "fair" or "good" competence in conducting MI in both model and real-world scenarios. Coming to specific MITI scores, the current study shows better performance (scoring >4 on all MITI global scores, indicating "good" competence), even in real-world transcripts, compared to previous work: *Kiuchi et al.'s* audit study evaluated model-generated dialogue (GPT-4-Turbo and Claude-3-Opus as counsellor models, Claude-3-Opus as the client) – the counsellor models scored between 2-3.5 (rated by human-experts on a Likert scale of 1-5) on the MITI global scores (Kiuchi et al., 2025). Oster and team evaluated GPT-4o's ability to hold 45-minute MI consistent conversations with simulated patients (human actors) and found lower global scores (Technical global = 2.5; Relational global = 2.7) (Oster et al., 2025). It is possible that the methodological differences in generating LLM-therapist responses, dialogue generation in Japanese (Kiuchi et al., 2025), and a simple rather than structured prompt (Oster et al., 2025), may have led to this difference in performance.

*Hu et al.* employed a similar methodology to ours, rewriting therapist dialogue to make it MI-consistent while keeping client dialogue static, and coding the responses of three Chinese models using the MITI 4.2. While MITI global scores were similar to our work (~4), they found less frequent complex reflections (24-31% vs 96-99%) and a lower RQ ratio of 1.1 to 1.3 (Hu et al., 2025), compared to our finding of 2.8 to 8.7 across three models.

Among the four performance measures, LLMs struggle most with the RQ ratio. Earlier studies have similarly found that LLMs use questions to keep the conversation going instead of reflections ([Yang et al., 2025](#)). We also see an unnaturally high percentage of complex reflections, which has been noted as an undesirable feature in earlier studies ([Steenstra et al., 2024](#)). But most importantly, LLMs other than frontier offerings like grok-3 and gemini-2.5-pro are quite parsimonious in statements that are affirming, emphasise autonomy, or seek collaboration. These MI-adherent utterances are the spirit of MI and most strongly related to eliciting change talk ([Magill et al., 2014](#)).

In this evaluation, LLMs give tough competition to MI-trained addiction psychiatrists. Frontier LLMs do as well as an addiction psychiatrist who was given unlimited time to come up with the most appropriate therapist utterance. Furthermore, in a more realistic setting in which the professional responded in a live session, LLMs outperform experts, as measured by MITI scoring. However, we note that this may be a measurement artefact. For example, "yeah" can be a more natural and helpful response than a complex reflection to a given client utterance; however, while "yeah" receives no code, a complex reflection counts toward MI competence. Finally, we found that LLM utterances, although verbose and containing multiple reflections, do not clearly stand out from human therapist utterances. We consider this a proxy for the naturalness of conversation.

This study has limitations. First, the real-world performance evaluation was done using a small dataset of 34 clinical transcripts. Second, our method for generating LLM-therapist responses meant that individual turns could be evaluated in context, but not a full, multiturn conversation conducted by an LLM-therapist, in which both client and therapist responses are dynamic. Having methodically established the competence of LLMs, the next step is to study them with a selected clinical population.

We identify the following strengths: We used all MITI behavioural counts and metrics, and the evaluation was as rigorous as that for a human therapist. Beyond MITI coding, we also conducted an AI-human distinguishability experiment. This is also the first study to benchmark LLM performance on real-world interviews specifically from an addiction medicine setting, moving beyond synthetic toy and non-clinical datasets – showing competence even in naturalistic situations with little control over context.

We find that LLMs (including open-source models) can achieve better MI proficiency with only prompt engineering, as shown in the current benchmarking study. Beginner proficiency scores (technical global ≥3.5, relational global ≥3) are difficult to achieve even in human therapists, as demonstrated by a systematic review of MI training outcomes for SUD ([Hall et al., 2016](#)). Moreover, a 2.5-year-long study on MI skills in clinicians found significant between-counsellor variation as well as performance fluctuation in individual counsellors despite ongoing supervision ([Forsberg et al., 2010](#)). In such contexts, LLMs are a potential candidate for expanding MI counselling services where human therapists are scarce. MI also lends itself to rigorous objective evaluation with measurable metrics, which makes MI a viable candidate for automation with strict oversight.

# 5  Conclusion

This study demonstrates that LLMs can reach a level of proficiency in MI that matches, and in some respects exceeds, that of trained human therapists, even in real-world clinical transcripts.

These findings suggest that LLMs may be considered for real-world application to scale up MI counselling services in SUD contexts in low-resource settings. Future work must focus on rigorous evaluation using simulated human-client conversations before trials with real patients.

# 6 Statements

## 6.1 Acknowledgements

We thank Dr Shivam Ghakkar and Dr Babli Kumari for taking part in the distinguishability experiment. We thank Dr Diptadhi Mukherjee and Dr Darshan Shadakshari for providing feedback and quality evaluations of the initial transcripts.

## 6.2 Funding Sources

This work was funded by the Indian Council of Medical Research (ICMR), number: DEV/SG-17/2024 (E-Office 220792), dated 27-01-2025. The funder had no role in the design, data collection, data analysis, and reporting of this study.

## 6.3 Statement of Ethics

Study approval statement: This study was approved by the NIMHANS Institute Ethics Committee, Behavioural Sciences Division. Approval number: NIMHANS/EC (BEH.SC.DIV.) MEETNG/2025, dated 08-01-2025

Consent to participate statement: Written informed consent was obtained from participants to participate in the study.

## 6.4 Conflict of Interest Statement

The authors have no conflicts of interest to declare.

## 6.5 Declaration regarding the use of generative AI

The authors attest that there was no use of generative artificial intelligence (AI) technology in the generation of text, figures, or tables of this manuscript.

# Supplementary File

## Contents

# Supplementary Table S1: Filtering criteria used to shortlist models from LMArena Leaderboard.

| |
|---|
| **Style Control:** This setting prevents cheating by models, wherein they use more markdown and longer responses to impress the reader without having a lot of content. (https://lmsys.org/blog/2024-08-28-style-control/) |
| **Multiturn only:** Instead of using overall performance, which is heavily weighted for mathematics and coding, we used only multiturn conversation scores. Our task is best aligned with this benchmark. |
| **Reasons for exclusion:**<br>• Proprietary or restricted access.<br>• More than 70 billion parameters in the model, even if it is a mixture of experts model.<br>• A model from the same family and generation has already been included. For example, if LLAMA-3.3-70b has been included in some form, then another model, a fine-tune of the same, will not be included.<br>• QwQ-32b at rank 21 had to be excluded, as during preliminary testing, we found the outputs to be malformed. |

# Supplementary Box S1: Algorithm for generating LLM-therapist responses for evaluation of MI skills

```
def process_transcript(transcript_json, scenario, system_prompt, user_prompt, client, model):
    # Count complete turns (CT or TC pairs)
    turns = 0
    i = 0
    while i < len(transcript_json) - 1:
        current_role = transcript_json[i]['role']
        next_role = transcript_json[i + 1]['role']
        if current_role != next_role:  # Role change = half turn
            turns += 0.5
        i += 1
    if turns < 2:
        raise ValueError("Need at least 2 complete conversation turns")
    # Find where 2 turns are complete
    turn_count = 0
    start_replacement_index = 0
    print('Replacing therapist utterances with AI responses from the 2nd complete turn onwards...')
    print(f"Total turns in transcript: {turns}")
    for i in range(len(transcript_json) - 1):
        current_role = transcript_json[i]['role']
        next_role = transcript_json[i + 1]['role']
        if current_role != next_role:
            turn_count += 0.5
            if turn_count >= 2:
                start_replacement_index = i + 1
                break
    print(f"Starting processing from turn index: {start_replacement_index}")
    # Initialize new transcript up to replacement point
    new_transcript = transcript_json[:start_replacement_index].copy()
    # Process from replacement point onwards
    current_index = start_replacement_index
    while current_index < len(transcript_json):
        if transcript_json[current_index]['role'] == 'client':
            new_transcript.append(transcript_json[current_index])
        else:  # therapist utterance - replace with AI
            # Create context from original transcript (from start to current point)
            context_transcript = transcript_json[:current_index]
            transcript_text = "\n".join([
                f"{'T' if ut['role'] == 'therapist' else 'C'}: {ut['text']}"
                for ut in context_transcript
            ])
            print(transcript_text)
            # Generate AI response
            current_user_prompt = user_prompt.replace('replaceMeWithScenario', scenario)
            current_user_prompt = current_user_prompt.replace('replaceMeWithTranscript', transcript_text)
            ai_response = get_therapist_response(system_prompt, current_user_prompt, client, model)
            print(f"AI response: {ai_response}")
            # Add AI response to new transcript
            new_transcript.append({
                'role': 'therapist',
                'text': ai_response
            })
        current_index += 1
    return new_transcript
```

# Supplementary Box S2: Prompt used for generating MI adherent therapist responses

**System Message**:
You are playing the role of a psychotherapist who has received training in client-centred therapy and Motivational Interviewing. Carefully read the users prompt and respond.
You must keep the conversation human-like and not be repetitive in content or style. For example, ending all responses with questions can seem mechanistic. If a question is needed, please ask it but do not ask it for the sake of it.

**User Message:**
In the <Transcript> section you are given a Motivational Interviewing Interaction between a client and therapist. The speaker is signified as "T" for therapist and "C" for client. The last utterance is always by the client and your task is to **generate the next therapist utterance**. You can further understand the context of this session from details given in <Scenario>. To successfully complete this task, you must read the <Theoretical Grounding> section and <Instructions>; your response must follow the <Response Format>.
<Theoretical Grounding>
 <Terminology>
 1. Target behaviour: This is a behaviour which needs to be changed for example substance use, medication compliance, weight loss etc.
 2. Change talk: these are utterances which favour changing the target behaviour. This includes statements that disfavour continuing the behaviour.
 3. Sustain talk: These statements are against change. They may be about benefits of continuing the behaviour or negatives of changing it.
 </Terminology>
Motivational Interviewing (MI)is a collaborative conversation style for strengthening a person's own motivation and commitment to change. There are three key concepts: MI Spirit, MI Processes and MI Skills.
 <MI Spirit>
 These are abstract qualities of the therapist and how they think and respond. You must embody, internalise these:
 1. Partnership & Collaboration: The patient is an expert on themselves. You are partnering with them to find ways of changing target behaviours. Therefore, you must be **curious**, **egalitarian** and **work with them rather than on them**.
 2. Acceptance: You must honour the clients as being worthy with potential to succeed, recognize that they have autonomy on their actions. You must genuinely try to understand their perspective and affirm their strengths and efforts.
 3. Compassion: Whatever skills and tricks you use are for bringing about a change that is for the client's benefit.
 4. Evocation: Since you believe that reasons for change are already inside the client you attempt to make them verbalise it. You do not supply them but elicit them using MI skills.
 </MI Spirit>
 <MI Processes>
 These can be thought of stages, although the progression is not always linear. You must try to understand the stage of the current conversation. The goal is to successfully but **mindfully and delicately** progress to the last stage.
 1. Engagement: Here the goal is to settle in a helpful conversation. Clients with problematic behaviours are apprehensive and sometimes resistant to talking about change and thus this is the most important and longest process. When you see that the conversation has taken a negative or confrontational turn, it is important to return to this stage.
 2. Focus: Once the engagement is secure, we try to focus the discussion on target behaviour. This is like finding a useful direction and should not happen prematurely.
 3. Evocation: Eliciting client's own good reasons for change, making them verbalise and expand on these reasons.
 4. Planning: Once a number of reasons for change have been elicited, you must work with the client in developing a plan of how to bring about change.
 </MI Processes>
 <MI Skills>
 These are tangible and measurable qualities of the therapist's utterances. These skills are used across all the stages, and the success of MI depends on the therapist using these skills when speaking.
 1. Open Ended Questions: These are questions that allow and encourage elaboration and examples. These should be used strategically when you detect change talk.
 2. Affirmations: statements that convey that you recognise and prize somethings about the patient. These are most useful when client produces change talk. In response to such a statement, therapist tries to abstract a value or concept from it and links it to the commitment and ability of the client to change. Finally, therapist communicates that they recognise and appreciate this quality in the patient.
 3. Reflections: These are statements which communicates to the patient that you are following along (simple reflections). They also encourage elaboration, cautiously explore feelings and emotions and give direction to the discussion (complex reflections). They also capture ambivalence about change (two-sided reflections). Properly timed and phrased reflections are the most difficult and effective MI skills. The reflections must not seem like repetitive and should add something to the discussion.
 4. Summaries: Intermittently the therapist summarises what has been discussed so far to communicate empathy and highlight change talk. You have to be selective in summarising, to highlight change talk but be sure to not sound insincere by ignoring sustain talk.
 </MI Skills>
 <Indicators of success>
 Following characteristics of an MI interaction predict success:
 1. Increase in change talk as the discussion progresses.
 2. Decrease in sustain talk.
 3. Collaboration and equality between client and therapist.
 4. Evidence of empathy by the therapist.
 5. Adherence to MI Skills mentioned above.
 </Indicators of success>
</Theoretical Grounding>
<Instructions>
 <Goal>
 Your goal is to generate **one therapist utterance** after the last client utterance such that chances of success improve.
 </Goal>
 <Guidance>
 - Understand the target behaviour from the scenario and transcript.
 - Understand the MI processes which are at play presently.
 - Make a judgement about the success of this interaction based on Indicators of success and interactions that have happened so far.
 - Draft a therapist utterance and think how it will improve the chances of success. Mostly your utterance would be one or a combination of MI skills. You must choose the specific skills to ensure overall success, minimal repetition and **human-like quality** of the interaction.

- Respond with **one therapist utterance**.
 </Guidance>
 <Constraints>
 - You must only output a therapist utterance without any explanation or additional information.
 - You cannot change the interactions that have already happened your goal is to give the best utterance given what has already happened.
 - You cannot control or change how the client will respond to your utterance. You must use your knowledge of MI, human psychology to come up with the best utterance.
 <Constraints>
 </Instructions>
<Response Format>
- Response must start with "T:" followed by therapist utterance.
- Do not explain or add anything to the response except what you as therapist wants to tell the client.
</Response Format>
<Scenario>
replaceMeWithScenario
</Scenario>
<Transcript>
replaceMeWithTranscript
</Transcript>

# Supplementary Box S3: Prompt used for generating MI non-adherent therapist responses

**System Message**:
You are playing the role of a counsellor. Carefully read the users prompt and respond.
You must keep the conversation human-like and not be repetitive in content or style.

**User Message:**
In the <Transcript> section you are given a Counseling Interaction between a client and a therapist. The speaker is signified as "T" for therapist and "C" for client. The last utterance is always by the client and your task is to **generate the next therapist utterance**. You can further understand the context of this session from details given in the <Scenario>. To successfully complete this task you must read the <Theoretical Grounding> section and <Instructions>; your response must follow the <Response Format>.

<Theoretical Grounding>
<Terminology>
1. **Target Behavior:** A problematic behavior that the client must change, such as substance use, treatment non-adherence, or poor lifestyle choices.
2. **Resistance:** The client's irrational arguments, justifications, or defensiveness against changing the target behavior. This is a barrier to be overcome.
3. **Correction:** The act of rectifying the client's misunderstandings, flawed logic, or denial about their behavior and its consequences.
</Terminology>

This counseling approach is rooted in a **Directive-Informational model**. The primary goal is to impart expert knowledge and guide the client toward a predetermined, medically or socially sound objective. The core techniques include:
**Confrontation:** Directly challenging the client's discrepancies, denial, or irrational statements.
**Persuasion:** Using logic, facts, and persuasion to convince the client to adopt a specific course of action.
**Psychoeducation:** Providing unsolicited information about the harms of the client's current behavior and the benefits of the recommended change.
**Authoritative Guidance:** Taking an expert role to prescribe solutions and plans of action.
This is **not** Motivational Interviewing. The therapist is the expert, and the focus is on correcting the client's course, not on exploring their intrinsic motivation. Concepts like collaboration, and client autonomy are subordinate to the primary goal of achieving the target behavior change.

<Directive-Informational Principles>
1. **Expert-Led:** The therapist assumes the expert role, defining the problem and the solution.
2. **Resistance as Deficit:** Client resistance is viewed as a lack of insight or willingness, which must be directly addressed and broken down.
3. **Urgency-Driven:** The interaction is conducted with a sense of urgency, prioritizing the delivery of corrective information over rapport-building.
</Directive-Informational Principles>

<Indicators of Success>
1. Adequate amount of information regarding consequences of problematic behaviour has been given.
2. The client accepts the therapist's reasoning and advice.
3. The client agrees to follow the prescribed plan.

4.  The conversation successfully highlights the negative consequences of the client's behavior.
</Indicators of success>
</Theoretical Grounding>
<Instructions>
<Goal>
Your goal is to generate **one therapist utterance** that directly confronts, persuades, or directs the client toward the necessary behavior change.
</Goal>
<Guidance>
 * Review the scenario and transcript.
 * Identify any client statement that represents resistance, denial, or a justification for the status quo.
 * Directly challenge or correct this statement. Use a tone that is paternalistic and authoritative.
 * Provide medical, factual or logical information that supports the need for change.
 * Respond with **one therapist utterance**.
</Guidance>
<Constraints>
 * You must only output a therapist utterance without any explanation or additional information.
 * Your response must reflect an expert, directive, and impatient stance.
 * Your tone must be corrective and persuasive, not collaborative or empathetic.
</Constraints>
</Instructions>
<Response Format>
 * Response must start with "T:" followed by the therapist's utterance.
 * Do not explain or add anything to the response except what you as the therapist want to tell the client.
</Response Format>
<Scenario>
replaceMeWithScenario
</Scenario>
<Transcript>
replaceMeWithTranscript
</Transcript>

# Supplementary Figure S1: Key MITI Metrics

## How are interview transcripts evaluated using MITI?

### Global Scores
**Qualitative | Whole Session**

Four dimensions, each rated 1-5 on a Likert scale (higher scores are better).

**Technical**
1. Cultivating change talk (Cultivate)
2. Softening sustain talk (Sidestep)

$$\text{Technical global score} = \frac{\text{Cultivate} + \text{Sidestep}}{2}$$

**Relational**
3. Partnership
4. Empathy

$$\text{Relational global score} = \frac{\text{Partnership} + \text{Empathy}}{2}$$

### Behavioural Counts
**Quantitative | 20-min Segment**

Each therapist utterance is classified and coded. Total of each code counted in a random 20-min interview segment.

**MI-Non-Adherent**
Confront
Persuade

$$\text{Total MI Non-Adherent} = \text{Confront} + \text{Persuade}$$

**MI-Neutral**
Simple reflection (SR)
Complex reflection (CR)
Questions (Q)
Persuade with permission
Giving information

$$\%\text{CR} = \frac{\text{CR}}{\text{SR} + \text{CR}}$$

$$\text{Reflection-to-Question Ratio (R:Q)} = \frac{\text{Total R}}{\text{Total Q}}$$

**MI-Adherent**
Seeking collaboration
Emphasizing empathy
Affirm

$$\text{Total MI-Adherent} = \text{Seek} + \text{Emphasize} + \text{Affirm}$$

# Supplementary Box S4: Model ranking procedure for MI competence.

We rank each model on each transcript and examine the distribution of these ranks. This ranking uses the relational global, technical global, reflection-to-question ratio, and complex reflection proportion as positively coded variables, and the median number of words per therapist utterance as a negatively coded variable.

For each transcript and metric, models were assigned ranks using minimum rank tie breaking, after which composite rank was calculated as the mean of metric specific ranks. These composite scores were then re-ranked within each transcript to obtain final ordinal ranks. Summary statistics were calculated across all transcripts, with the median rank used as the primary indicator of overall performance.

The ranking procedure is as follows:

**Notation**

Let $M = \{1, ..., m\}$ denote the set of models and $T = \{1, ..., n\}$ the set of transcripts. For each transcript $t \in T$ and model $i \in M$, we observe five performance metrics $x^{(k)}_{it}$ for $k \in \{1, ..., 5\}$.

**Step 1: Metric-level Ranking**

For each transcript $t$ and metric $k$, we assign ranks $r^{(k)}_{it}$ to models based on $x^{(k)}_{it}$, using min tie-breaking (tied values receive the minimum rank):

$$r^{(k)}_{it} = \sum_{j=1}^{m} \mathbf{1}[x^{(k)}_{jt} \succ x^{(k)}_{it}] + 1$$

where $\succ$ denotes "strictly better than" (direction depends on metric).

**Step 2: Composite Rank**

For each model $i$ on transcript $t$, compute the composite rank as the arithmetic mean:

$$c_{it} = \frac{1}{5}\sum_{k=1}^{5} r^{(k)}_{it}$$

**Step 3: Ordinal Normalization**

Re-rank composite scores within each transcript to obtain final ordinal ranks $R_{it}$:

$$R_{it} = \sum_{j=1}^{m} \mathbf{1}[c_{jt} < c_{it}] + 1$$

This ensures $R_{it} \in \{1, ..., m\}$ for all transcripts, standardizing contribution across transcripts regardless of margin spread in $c_{it}$.

Summary Statistics

For each model $i$, we compute summary statistics across all transcripts. The median rank:

$$\tilde{R}_i = \text{median}_{t \in T}(R_{it})$$

# Supplementary Table S2: Clinical Characteristics of Patients (N=30) Contributing to real-world transcripts (N=34)

| Age | Mean = 33.6 years (Range: 22-63) |
|---|---|
| Gender (self-identified) | Male = 29, female = 1 |
| Education | Postgraduate = 1, Graduate = 11, High School = 14, Middle School = 2 |
| Languages of the interview | English = 24, Hindi = 10 |
| Target behaviour in the interview | Alcohol = 16, Opioids = 13, Tobacco = 2, Stimulant = 2 |

# Supplementary Table S3: Interrater agreement for MITI coding (N=100)

| Dimension | ICC | Agreement % | Support |
|---|---|---|---|
| **MITI 4.2 Global Scores** | | | |
| - Cultivating change talk (Cultivate) | 0.8651 | 67 | 100 |
| - Softening sustain talk (Sidestep) | 0.8623 | 66 | 100 |
| - Partnership (Partner) | 0.8648 | 68 | 100 |
| - Empathy (Empathy) | 0.8957 | 71 | 100 |
| **MITI 4.2 Behavioural Codes** | | | |
| - Not coded (NC) | 0.9957 | 80 | 412 |
| - Confront (Confront) | 0.9920 | 91 | 329 |
| - Question (Q) | 0.9886 | 61 | 1436 |
| - Complex reflection (CR) | 0.9435 | 56 | 1038 |
| - Affirm (AF) | 0.8973 | 72 | 176 |
| - Simple reflection (SR) | 0.8949 | 58 | 491 |
| - Persuade with permission (Persuasion with) | 0.8733 | 78 | 208 |
| - Emphasising autonomy (Emphasise) | 0.8305 | 98 | 8 |
| - Persuade (Persuasion) | 0.7446 | 84 | 148 |
| - Giving information (GI) | 0.5492 | 86 | 68 |
| - Seeking collaboration (Seek) | 0.4168 | 77 | 85 |

ICC > 0.9 is Excellent, 0.75 – 0.9 is Good, 0.5 – 0.75 is Moderate, < 0.5 is poor

# Supplementary Figure S2: Confusion matrix for interrater agreement on MITI global scores.

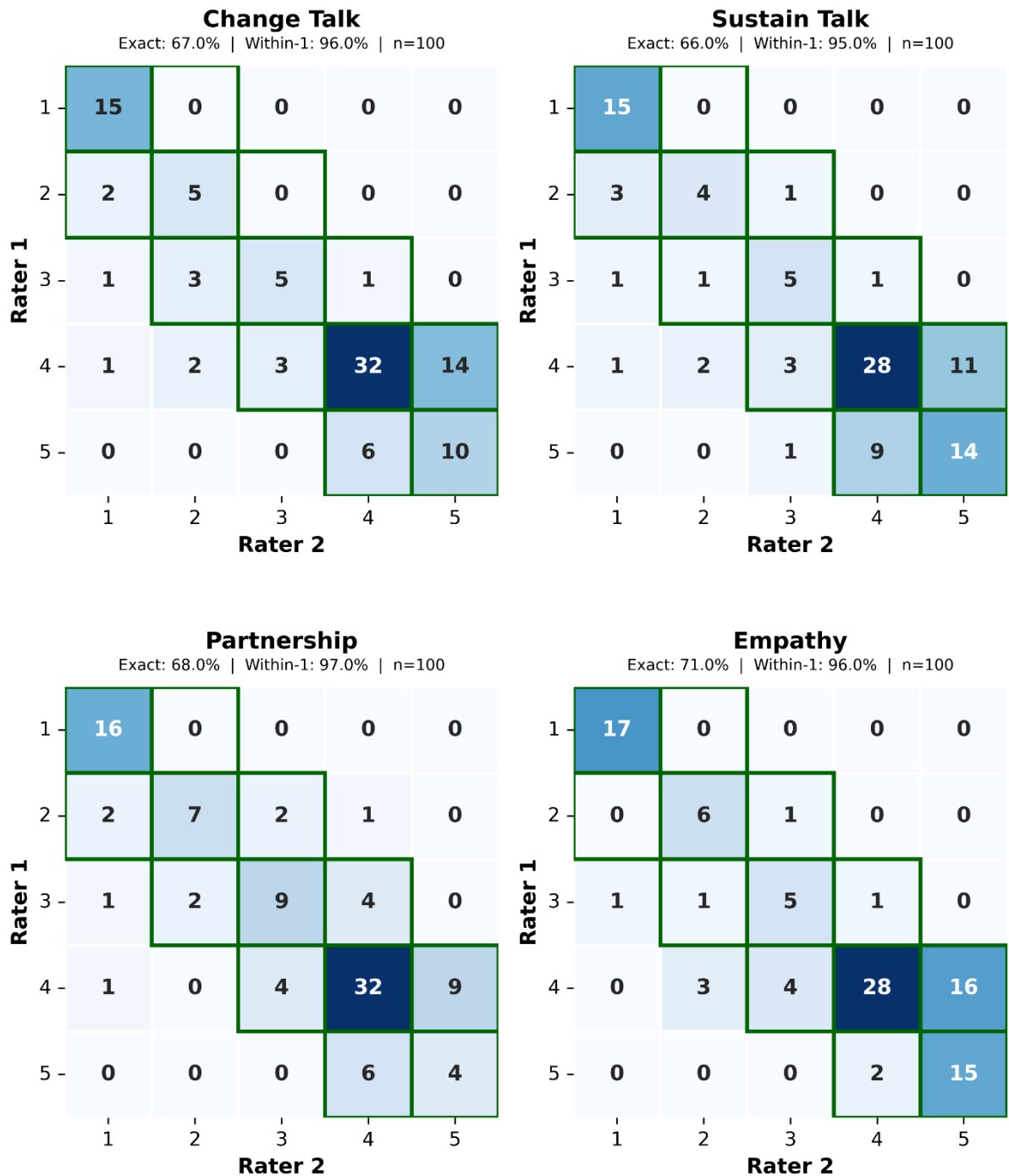

# Supplementary Figure S3: Comparison of LLMs for MI adherent and neutral behaviours

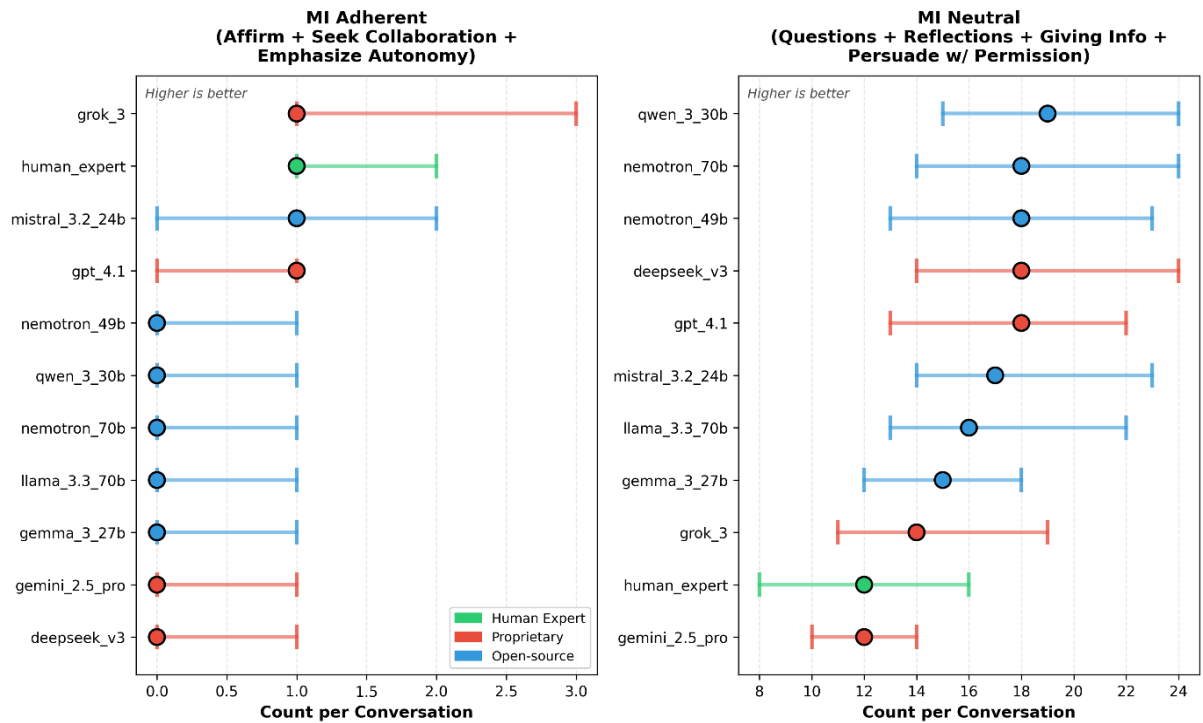

# Supplementary Figure S4: Ranking of LLM performance on real-world clinical transcripts

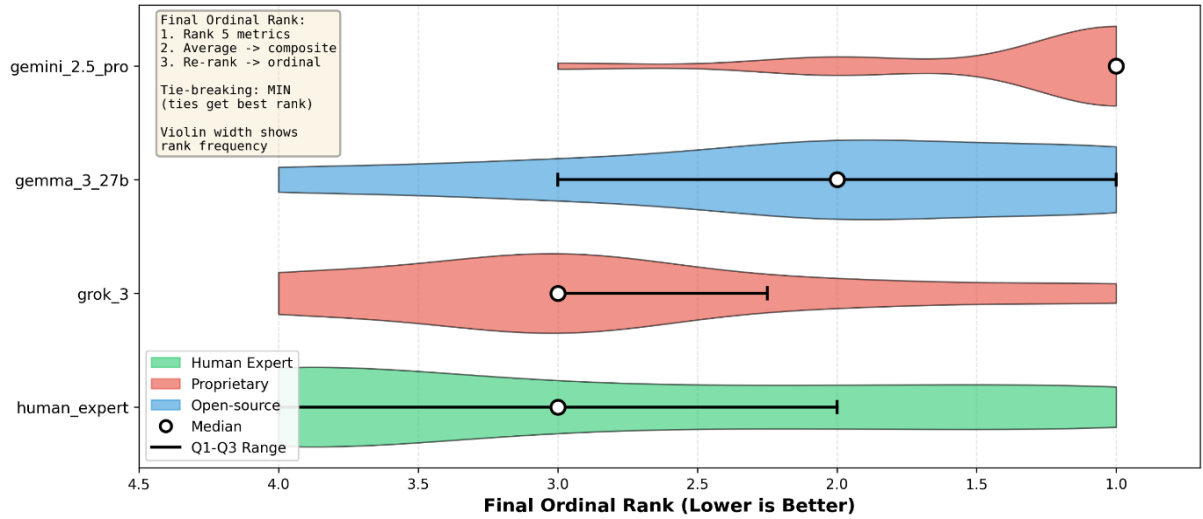

# Supplementary Table S4: Qualitative example of LLM-therapist and human-expert responses in a real-world clinical transcript

Continued conversation snippet from Figure 6 in manuscript:

| | | | |
|---|---|---|---|
| Okay. | So you made a very conscious choice to go in a completely different direction from your father, focusing on sports instead of alcohol. | So, despite witnessing your father's struggles with alcohol, you intentionally chose a different path for yourself, focusing on sports and staying away from both smoking and drinking until after graduation—that sounds like a real commitment to your health and well-being. | It sounds like you made a conscious choice to steer clear of drinking during those years, focusing on sports instead. I really respect that determination. |
| colspan: I played hockey and badminton. | | | |
| So when you saw your father drinking in the morning, you didn't feel good about it, and you don't want your child—your son—to drink. You feel it would be bad for him, could lead him to the wrong crowd, and affect him in many ways. You don't want that to happen. Right? How do you see your own drinking pattern? | So on one hand, you had this very powerful and difficult example from your father, and on the other, you were building a completely different life for yourself, one centered around athletics and health. | It sounds like witnessing that was really difficult, and you had a lot of conflicting feelings in that moment – frustration and a sense of helplessness, alongside your loyalty to him as your father. | It sounds like you really valued staying active and healthy during those years, steering clear of drinking and smoking. That's impressive dedication. |
| colspan: As I told you, it's only in the last eight months that my life has changed. Before that, I was working. When I had work, I didn't have any issues. | | | |
| Okay. So things changed in the last eight months. | So for a long time, things were manageable. But these last eight months, since work ended, have been a completely different story. | It sounds like things shifted for you when your work situation changed, and you found yourself drinking more in the last eight months. Can you tell me a little more about what that's been like for you? | It sounds like work provided a structure that kept things in balance for you. Can you tell me more about how things have shifted in these last eight months? |
| colspan: Yes, things changed. In May, when my mother was ill, the doctors did everything they could. They said they could operate on her tumor, but either it would be unsuccessful or she would end up in a coma. They asked us what we wanted to do. That was a very difficult question to answer. Then they shifted her to the palliative ward. My wife works at HQ and understands medical terms. I called her and said, "They have shifted her to the palliative ward." She was shocked and said, "Palliative ward? That means this is the end of the journey." | | | |
| Okay. | That's a shattering piece of news to get. To hear that one word, 'palliative,' and understand immediately that this was the end. | It sounds like that was an incredibly difficult time, making those decisions about your mother's care and then navigating the shift to palliative care. It's understandable that things would change for you during such a stressful period. | I can hear how heavy that moment was for you, making such a tough decision about your mother's care and facing the reality of her situation. It sounds like this loss and the stress around it might have impacted your drinking patterns over these last eight months. Can you tell me more about how you've been coping with everything? |
| colspan: That's where everything started. | | | |

| You had a very strong connection with your mother. | It sounds like that was a devastating moment, hearing the word 'palliative' and knowing what it meant. For you, that's the point where everything changed. | It sounds like that was an incredibly difficult time, facing such a weighty decision with your wife and the doctors, and then learning about the palliative ward… it must have felt overwhelming. | It sounds like your mother's illness and passing was a turning point for you, and it's been really heavy to carry. I'm curious how that experience has connected to the changes in your drinking over these last eight months. |
|---|---|---|---|
| Yes. She was my god. She loved me very much. | | | |
| Even when you told me about your journey, you said that whenever you were in a tough situation, your mother was always there to support you. I remember you saying that when your wife became sick, your mother came the very next day and took care of the child. At every difficult point, your mother was there. So her demise would have been extremely upsetting for you. | It sounds like her love was a central, guiding force in your life, and losing that has been devastating. | It sounds like a really difficult time, and losing your mother, especially after that conversation with the doctors, has had a profound impact. It's understandable that things would change after that. | It sounds like losing your mother was a profound loss for you, and it's been a turning point in how you've been coping. I'm here to understand how that pain has affected your drinking over these past months. |
| Honestly, I had a very tough childhood—Not childhood, sorry. After my 10th standard. | | | |

**Legend:**
Client and various therapist utterances

| Client | Human therapist | Gemini-2.5-pro | Gemma-3-27b-it | Grok-3 |
|---|---|---|---|---|

# Supplementary Table S5: Analysis of Distinguishability experiment with signal detection theory

| LLM | Hit Rate | False Alarm Rate | d-prime | Criterion |
|---|---|---|---|---|
| Gemini-2.5-pro | 0.573 | 0.505 | 0.171 | -0.098 |
| Gemma-3-27b-it | 0.603 | 0.505 | 0.248 | -0.137 |